\documentclass[letterpaper, 10 pt, conference]{ieeeconf}
\IEEEoverridecommandlockouts
\usepackage{amsmath,amsfonts}
\usepackage{algorithmic}
\usepackage{algorithm}
\usepackage{array}
\usepackage[caption=false,font=normalsize,labelfont=sf,textfont=sf]{subfig}
\usepackage{textcomp}
\usepackage{stfloats}
\usepackage{url}
\usepackage{verbatim}
\usepackage{graphicx}
\usepackage{cite}
\usepackage{array}
\usepackage{multirow}
\usepackage{xcolor}

\usepackage{amssymb} 
\usepackage{booktabs}
\begin{document}

\title{DASH: A \textbf{d}a Vinci \textbf{A}dapter for \textbf{S}erial-link and \textbf{H}umanoid Robots \\ 
as an Accessible Platform for Surgical Robotics Research}
\author{Sara Wickenhiser$^1$, Junrong Zhou$^1$, Zekai Liang$^2$, Lizzie Peiros$^2$, Liu Shanglei$^3$, \\Michael C. Yip$^2$ \IEEEmembership{Senior Member,~IEEE}
\thanks{$^1$Mechanical and Aerospace Engineering Department, University of California San Diego, La Jolla, CA 92093 USA {\tt\footnotesize\{swickenhiser, juz058\}@ucsd.edu}}%
\thanks{$^2$Electrical and Computer Engineering Department, University of California, San Diego, La Jolla, CA 92093 USA. {\tt\footnotesize\{z9liang, epeiros, m1yip\}@ucsd.edu}}
\thanks{$^3$ Department of Surgery, University of California San Diego, La Jolla, CA, USA {\tt\footnotesize\{s5liu@health.ucsd.edu\}}}
}

\maketitle

\begin{abstract}
Robotic minimally invasive surgery offers well-documented clinical benefits, but the cost and infrastructure requirements of purpose-built platforms limit access in rural and lower-resourced facilities. Recent work has teleoperated general-purpose robots for laparoscopic tasks and in vivo procedures, but relied on handheld instruments coupled through passive linkages rather than native robotic actuation. Instead, we adapt da Vinci Classic and Xi instruments onto general-purpose robots that can integrate in clinical workflows. We present DASH, a da Vinci Adapter for Serial-link and Humanoid platforms, consisting of two types of adapters that require no modification to the instruments themselves. Each adapter is compatible with many robotic platform load capacities, engages the instrument's native latch, and wirelessly identifies inserted tools to load instrument-specific kinematics and coupling matrices. Teleoperated experiments demonstrate the efficacy of DASH and its benefits for enabling research access to surgical robotic platforms.

\end{abstract}

\section{Introduction}

Minimally Invasive Surgery (MIS) offers patients smaller incisions, faster recovery times, and lower surgical risk compared to traditional open procedures \cite{reddy2023}. Today, the gold standard for robotic MIS is purpose-built surgical robotic platforms, such as the da Vinci Surgical System, which has been widely adopted across major hospitals worldwide. Although these highly specialized platforms have demonstrated exceptional performance and achieved widespread clinical adoption, most existing operating rooms were designed for manual operation, causing modern robotic platforms to require extra coordination for equipment positioning \cite{spacebarriers2022}. Typical surgical platforms have large physical footprints with dedicated infrastructure requirements \cite{roomsize2021}, limiting their accessibility, particularly in resource-constrained settings where an entire suite dedicated to one platform may be infeasible.

\begin{figure}[t]
  \centering \includegraphics[trim={0.0cm 3cm 0cm 3cm}, clip, width=0.98\linewidth]{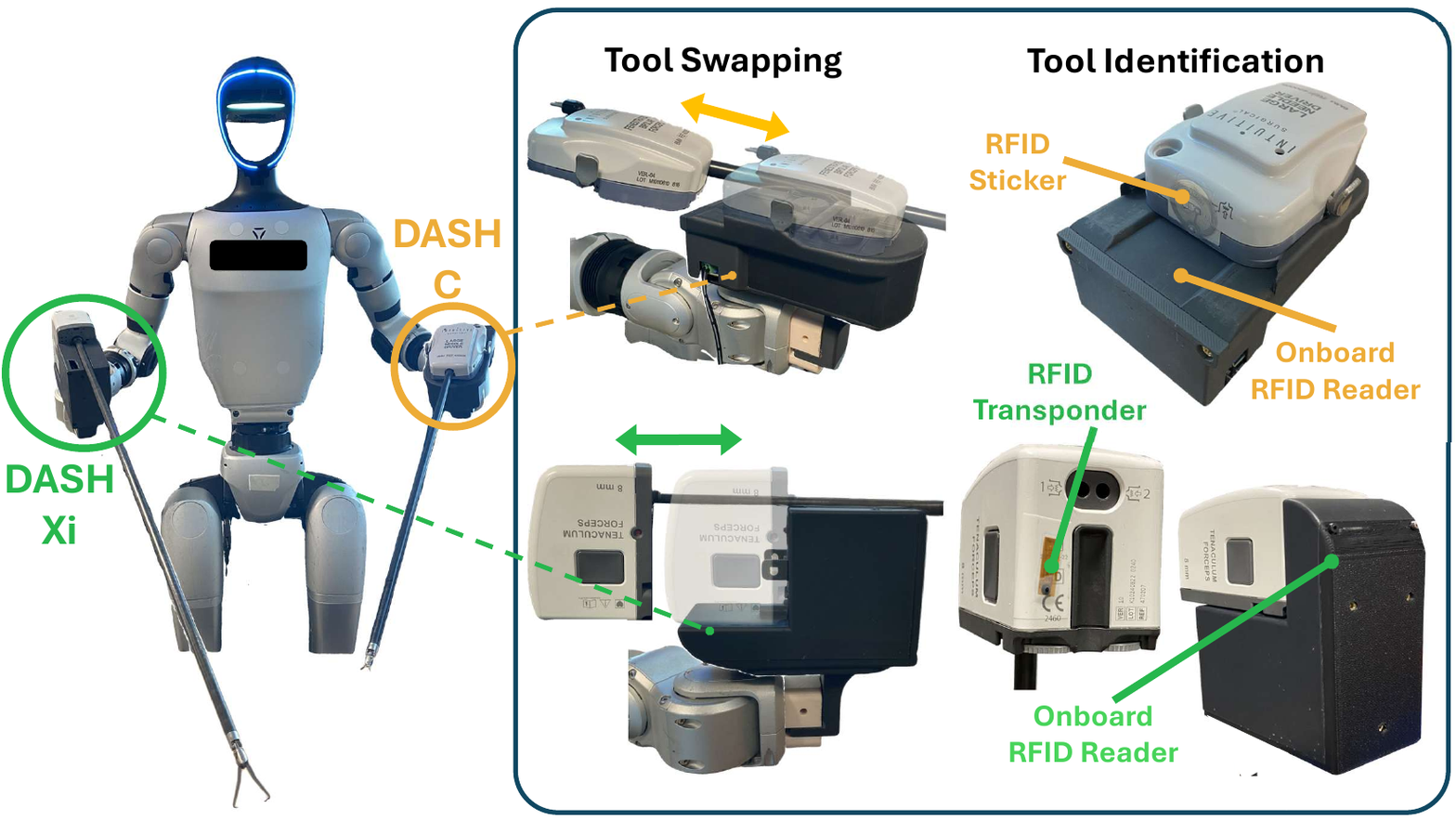}
  \caption{The da Vinci Adapter for Serial-link and Humanoid platforms (DASH) adapts da Vinci surgical instruments onto compact, versatile platforms such as a humanoid robot. By decoupling these instruments from the da Vinci system, DASH provides a platform that can substantially reduce the infrastructure barrier to highly precise robotic minimally invasive surgery.}
  \label{fig:coverFig}
\end{figure}

The core challenge lies not in the surgical procedure itself, but in the infrastructure required to perform it.  Recently, general-purpose robotic platforms have been explored as a potential pathway towards more accessible laparoscopic surgery. Humanoid robots \cite{humanoid2026,hum_to_human, general_humanoid} offer unique advanta
ges with their human-like morphology, allowing them to operate within existing surgical infrastructure and workflows. Furthermore, general purpose humanoid robots have recently emerged as viable platforms for laparoscopic surgery \cite{lapsurgie}, including teleoperated in vivo procedures \cite{liang2026}. Together, these studies highlight the potential of general-purpose humanoid systems to broaden access to high-quality surgical care. Beyond surgical settings, humanoid robots have been applied to support patient interaction \cite{patient_interaction}, bedside assistance \cite{bedside}, and hospital logistics \cite{logistics}. Using a versatile platform such as a humanoid robot can reduce the overall infrastructural requirement for robotic surgery, potentially decreasing the need for a dedicated operating suite for robotic surgical platforms. A platform that can contribute across multiple clinical workflows, rather than serving a single purpose, may improve resource utilization while reducing capital, storage, and operational demands, particularly in resource-constrained settings. 

Enabling humanoid robots and serial-link platforms to use general surgical robotic end effectors, such as the da Vinci surgical tools, facilitates robotic platforms to have full actuated control over position and orientation of surgical instruments. This integration makes general-purpose robotic platforms more versatile and easier for healthcare systems to incorporate into existing workflows, helping address workforce shortages. This work introduces two separate adapters, one designed for the Classic and one for the Xi da Vinci surgical instruments, allowing these instruments to be transferred from the da Vinci platform onto smaller, more versatile robotic systems. 

DASH utilizes the latching mechanism on board the tool itself, wirelessly identifies the inserted tools, adapts to its coupling matrix, and fits on both robotic humanoid and serial-link platforms, distinguishing it from existing surgical adapters. Intraoperative instrument exchange is a routine and important aspect of any surgical operation. Previous adapters, including MOPS \cite{mops2021}, enable the use of da Vinci surgical tools for research but do not support wireless identification or automatic adaptation for different instruments, nor do they leverage the native latching mechanisms of Classic and Xi tools. In contrast, our platform places particular emphasis on compatibility with humanoid robots, positioning DASH as a step toward a general humanoid surgical assistant that can be deployed in hospitals.

Both DASH adapters support tool interchangeability, allowing multiple surgical instruments to be swapped in and out while preserving the procedural versatility required across different surgical tasks. By decoupling the high precision instruments from the proprietary system, the infrastructure burden of offering robotic minimally invasive surgery is reduced. In this work, our contributions are threefold: 
\begin{itemize}
    \item The design and fabrication of separate da Vinci Classic and Xi tool adapters compatible with serial-link and humanoid platforms
    \item A wireless tool-identification pipeline that loads instrument-specific kinematics and calibration on insertion
    \item 
   Extensive quantitative evaluations of DASH have been conducted on a humanoid platform against the previously developed humanoid laparoscopic system~\cite{liang2026}, demonstrating substantial improvements in both operational responsiveness and RCM control precision.
\end{itemize} 

\section{Related Works}

\subsection{Conventional systems for surgical robotics research}
The Da Vinci Surgical System drastically advanced MIS, increasing precision, lowering recovery times,  reducing surgeon fatigue \cite{reddy2023}, and standardizing readily available MIS tools. The Da Vinci Research Kit (dVRK) \cite{dvrk} is often used in research labs to continue advancing surgical robotics \cite{dvrk2021}. The dVRK can perform tasks like cutting \cite{sim2realcutting2025}, suturing \cite{autosuture2025} \cite{mpcsurture2024}, removing debris\cite{irrigation2025}, and cauterizing \cite{xiao2025}. The tools the dVRK uses are readily available, making these tools accessible and standardized. The ease of access motivates the need for adapters such as DASH that enable other robotic platforms to use highly precise surgical equipment. 

Johnson \& Johnson recently released a more compact surgical platform called OTTAVA \cite{ottava2026}. This surgical platform is integrated to the operating table, making it more compact and transportable. DASH was designed to enable humanoid robots to bring this level of precision to surgical settings, with the added benefit of having potential to be even more versatile than a standalone surgical platform.

\vspace{-3mm}

\subsection{Humanoid robots}

Humanoid robots are versatile and have been shown to assist with critical aspects of human medical assessments \cite{atar2025}. Beyond this, humanoid robots have also been shown to succeed with manual handheld laparoscopic tools in teleoperated in-vivo surgeries \cite{liang2026}. While impressive, this work can be expanded by replacing the handheld laparoscopic tools with surgical tools that can be controlled completely through actuation, like the da Vinci surgical tools. Additionally, humanoid robots have been used as assistants in a cadaveric sphenoidectomy \cite{cho2026}. 

\subsection{Serial-link robots}
Serial-link robots such as the Franka Emika Panda robot are used throughout hospitals for various medical tasks including MIS. An example of one such platform is the Mazor X, which was shown to reduce postoperative hospital stays by almost 34\% \cite{mazor2025}.

There are many types of laparoscopic MIS tool mounts, enabling robots, like humanoid and serial manipulators, to perform surgical tasks \cite{atar2025}. Other platforms utilize the tools end-effector to 'swap' between tool tips rather than changing out the whole instrument \cite{mckinley2016}. Few platforms exist that enable the swapping of whole tools, but one such platform is MOPS \cite{mops2021}. While MOPS enables the swapping of tools, it fails to wirelessly identify which tool is inserted and adapt to its kinematic constraints, a crucial step for seamless operational use. 

\vspace{-3mm}

\section{System Design}


DASH C and DASH Xi are adapters which enable humanoid and serial-link platforms to actuate and interchange da Vinci Classic and Xi surgical tools. DASH can wirelessly identify the attached tools and update the instrument-specific kinematics and calibration. Both adapters weigh less than 0.5 kg with tool inserted and are compatible with load capacities of many humanoid and serial manipulators.

\vspace{-1.5mm}

\subsection{Requirements} 
DASH must connect to and control the tendon driven actuation systems on da Vinci Classic and Xi instruments and provide sufficient mechanical power appropriate for typical tasks. It must also include a wireless system to identify which tool is in use to apply correct joint limits and kinematics. The adapter must stay within a weight range, enabling it to be compatible with most commercially used manipulators like humanoid robots such as the Unitree G1 and serial-link robots such as the Franka Emika Panda or KUKA robot. Each adapter must have a mounting hole pattern that allows connection to any robot-platform interface. Additionally it would be desirable to enable DASH to be mounted at any desired angle. A summary of the requirements is provided in Table \ref{tab:dr}.

\begin{table}[H]
\vspace{2mm}
\centering
\caption{Design Requirements}
\label{tab:dr}
\begin{tabular}{m{3.15cm} m{2.15cm} m{2.15cm}}
  \toprule
  \textbf{Item} & \textbf{DASH C } & \textbf{DASH Xi} \\
  \midrule
  Tool Compatibility & Classic & Xi\\
  \midrule
  Mass (Including Tool) & $<$ 0.5 kg & $<$ 0.5 kg \\
  \midrule
  Degrees of Freedom & 4 & 4-5 \\
  \midrule
  Controls & EE Orientation & EE Orientation \\
  \midrule
  Actuator Mech. Power & 2 W / motor& 2 W / motor\\
  \bottomrule
\end{tabular}
\end{table}

\begin{figure*}[t]
  \centering \includegraphics[trim={0cm 5cm 2cm 6.2cm}, clip, width=0.98\linewidth]{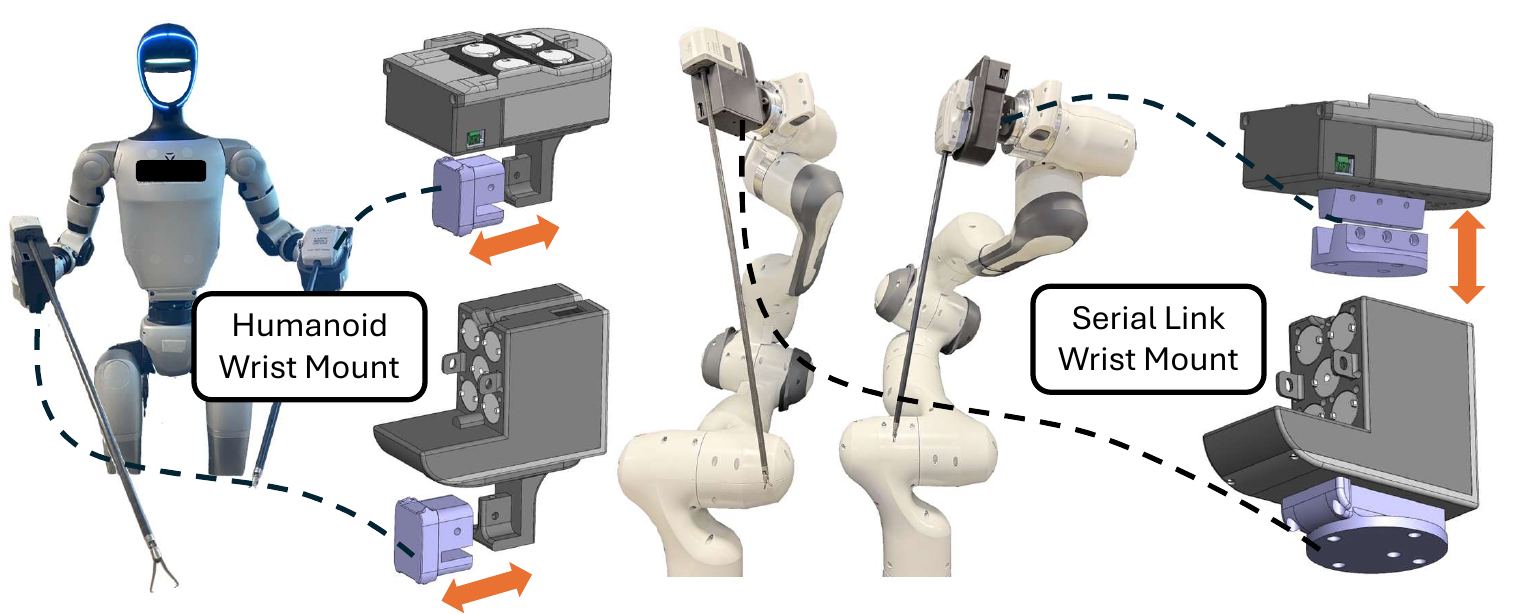}
  \caption{The humanoid and serial manipulators operating with da Vinci Xi or Classic surgical tools (real). Both adapters can easily mount to the platforms through interchangeable mounts that connect to the underside of each DASH type (shown in CAD).}
  \label{fig:surgie}
\end{figure*}

\begin{figure}[t]
  \centering \includegraphics[width=0.9\linewidth]{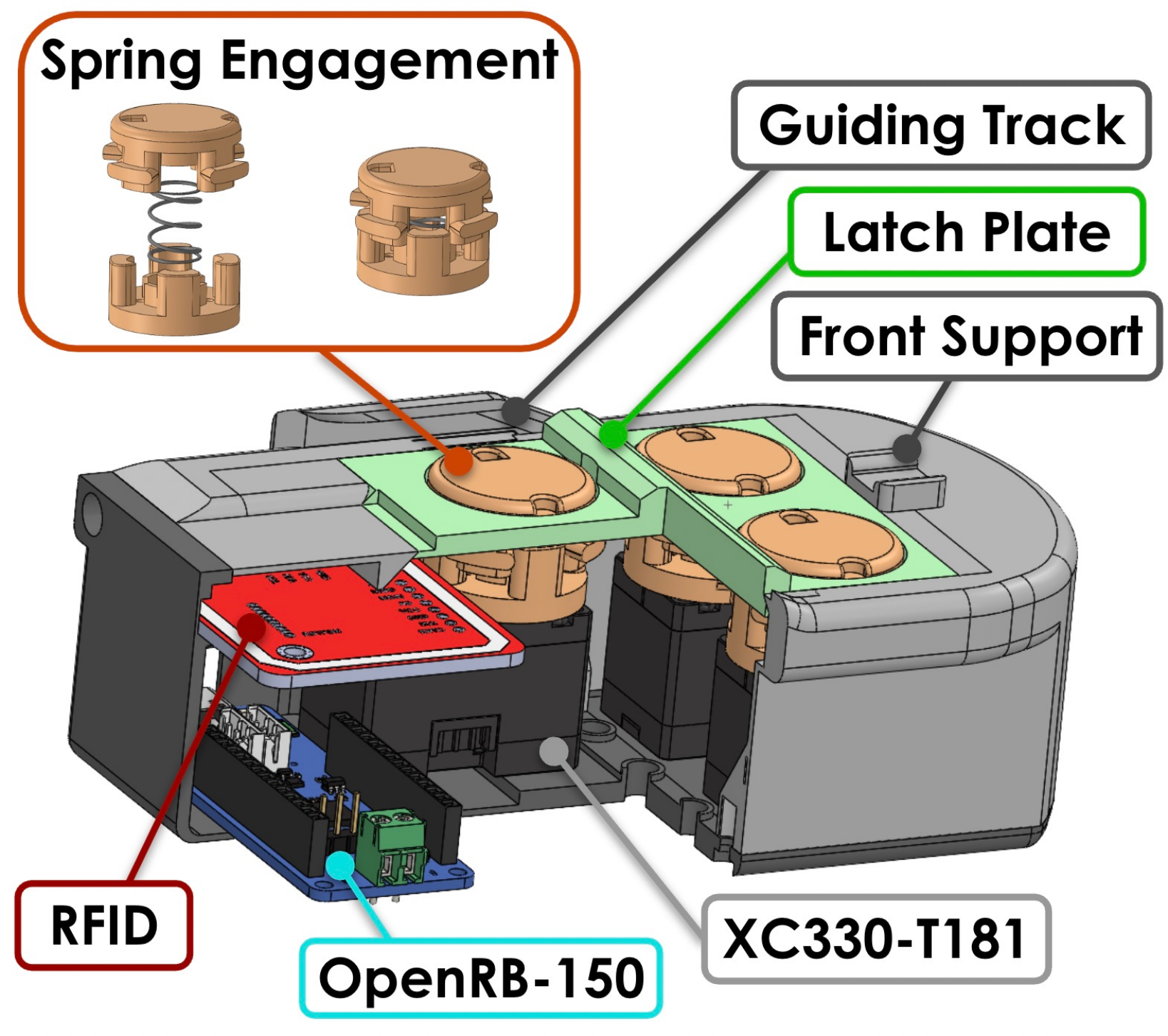}
  \caption{Overview of the DASH C's core features, including the guiding track, front support, spring engagement, RFID reader, motors, and controller. The DASH C is a compact and effective adapter that can connect with all the Classic tool instruments.}
  \label{fig:siCAD}
\end{figure}

\begin{figure}[t]
  \centering \includegraphics[width=0.9\linewidth]{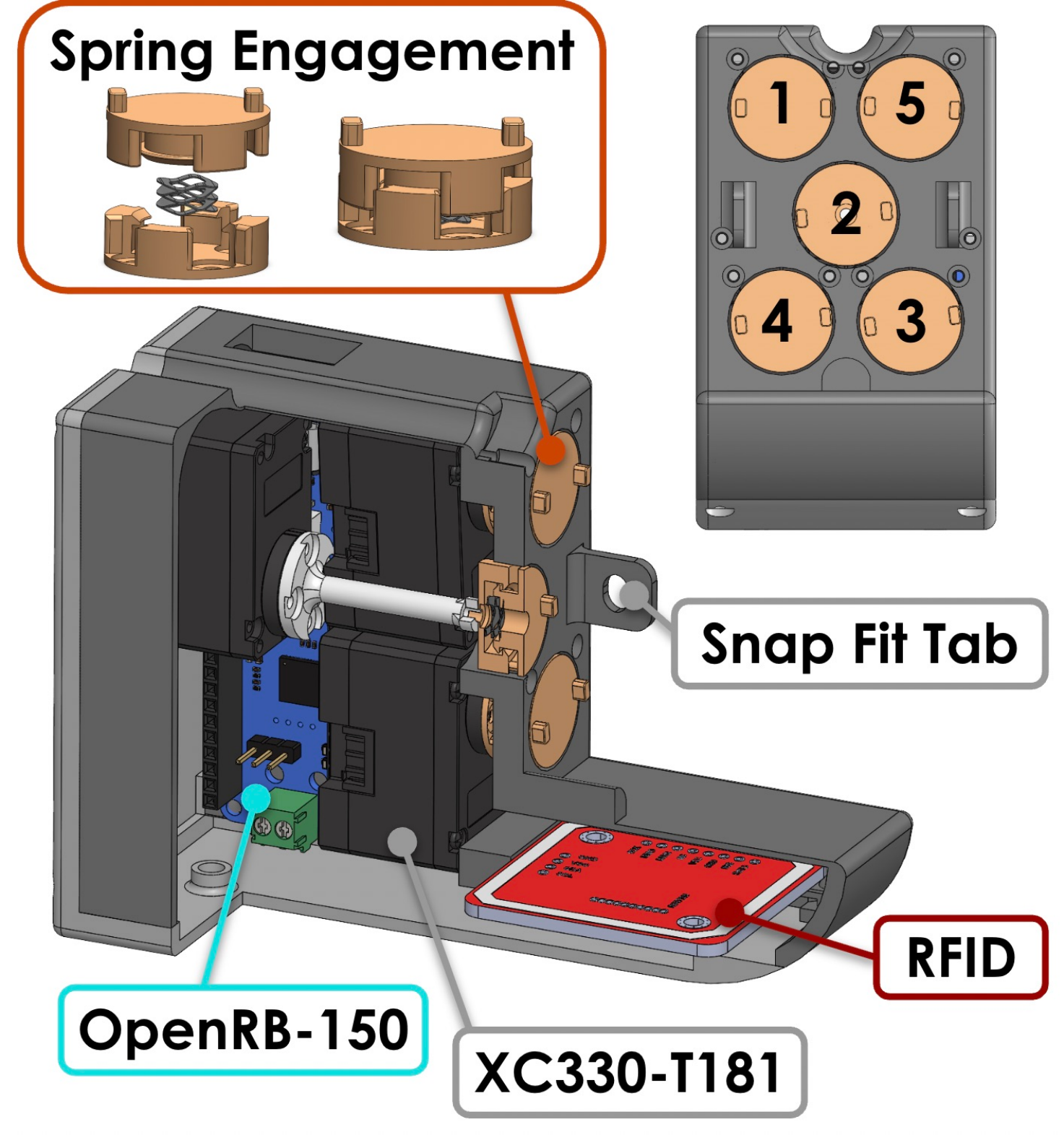}
  \caption{Overview of DASH Xi's core features, including the snap fit tab, spring engagement, RFID reader, motors, and controller. DASH Xi can connect to all Xi tool instruments. The front view in the top right shows the motor IDs.}
  \label{fig:xiCAD}
\end{figure}


\subsection{Robot-Platform Interface}

DASH is designed as an assembly of parts, one of which is the wrist interface, which aligns to the mounting point of the chosen robotic platform. These parts can be used interchangeably without modifying the electronics. While this work shows only one orientation, variations in this adapter plate can maximize workspace for each unique operation.

The interfaces must satisfy two independent interface constraints: (1) interfacing with DASH on one side and (2) mounting to  the robot end effector on the other. The interfaces used in this work are for mounting to the Unitree G1 humanoid robot, and the Franka Emika Panda serial manipulator shown in Fig. \ref{fig:surgie}. These platforms were chosen due to their frequent usage in research settings, demonstrating the versatility of this design. 

\subsection{Mechanical Design: Classic Tool}

DASH C, seen in Fig. \ref{fig:siCAD}, is composed of three key design features: the alignment features, the spring engagement, and the motor pack, referring to the general housing where the motors are mounted.

The alignment features include the guiding track and front support, which constrains lateral motion by engaging with the built-in features of the da Vinci Classic surgical tools. This configuration guides the instrument during insertion and minimizes undesired motion during operation.

The spring engagement secures the da Vinci tool and operates with the sliding insertion mechanism connecting the motors to the Classic tool actuation system to transmit torque. During insertion, the tool compresses the springs by pressing down the latch plate. Once fully inserted, the springs provide an upward force on the latch plate to lock the tool in place. The spring engagement consists of an upper cap that supports the latch plate and a lower cap that restrains the spring, with a compression range of approximately 3 mm, providing upwards force. This force maintains secure engagement without hindering the ease of tool exchange. Pressing the tool’s built-in release mechanism depresses the latch plate and disengages the tool from DASH C. 


\subsection{Mechanical Design: Xi Tool}
The da Vinci Xi tool mount, seen in Fig. \ref{fig:xiCAD}, is composed of three design features: the spring engagement, the latching feature, and the motor pack, similarly referring to the overall electronics housing where the motors are mounted.

The spring engagement allows compression when the tool features are not properly aligned to prevent direct mechanical interference that could damage the tool. The upper cap is snap-fit into the lower cap and incorporates a 3 mm wave spring, providing up to 2 mm of compression. The spring engagement is directly mounted to motors 1, 3, 4, and 5 and is connected to motor 2 through a shaft, as shown in Fig. \ref{fig:xiCAD}. The entire spring assembly is resin-printed using Gray V5 material to achieve high accuracy and strength.

The latching features ensure tool engagement upon insertion into DASH Xi. The snap-fit tabs engage with the built-in latching feature of the da Vinci Xi tool. The outer casing is made with Tough 1500 Resin V2 to provide sufficient flexibility and strength to securely hold the tool.

\begin{figure*}[t] 
    \centering
    \includegraphics[trim={0.1cm 5.8cm 0.5cm 6cm}, clip, width=0.98\textwidth]{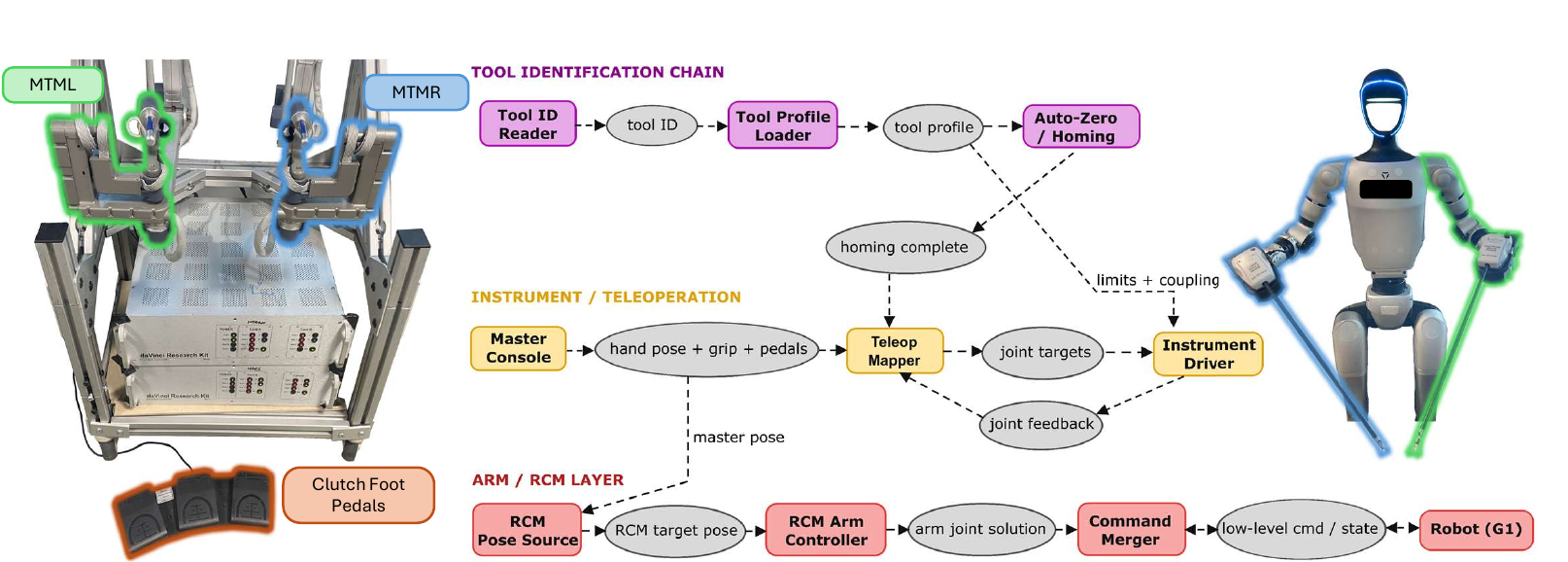}
    \caption{Humanoid surgical teleoperation system with DASH.
    After a tool is inserted, the ID is read and kinematics are updated.
    The Master Tool Manipulator (MTM), shown on the left, publishes target instrument poses, commanding the humanoid and DASH to position the end effector in the desired poses with the RCM constraint.}
    \label{fig:wide_image}
\end{figure*}

\subsection{Electronics}


The motor packs for both DASH adapters include an OpenRB-150 controller, a PN532 NFC RFID reader, and DYNAMIXEL XC330-T181-T motors which communicate over a TTL bus. The OpenRB-150 controls both the motors and RFID module, and the RFID board is positioned close to the surface for tool identification. Four motors actuate four degrees of freedom (DOF) of the Classic tool, and up to five motors are used in DASH Xi. 20 gauge wire was used to support the motors within both DASH types with 11.1~V input and max stall currents of 3.2~A and 4.0~A for DASH C and DASH Xi, respectively.

\subsection{Software/State Machine}

Tool changes were managed by a two-state control system that switched between searching for a tool ID and controlling the motors. When no tool was attached, the system remained in the searching state, continuously reading until a new tool was inserted and recognized.
The RFID reader identified the specific tool from RFID stickers affixed to the tool's back, and this identification was used to load the corresponding joint limits and coupling matrix. 
Upon insertion, an auto-zero routine rotated each motor $\pm$ 50$^\circ$ and back to zero, ensuring the spring engagement interlocked with each of the tool's joints, establishing a mechanical zero reference before teleoperation commands were accepted. 
Once the tool was zeroed and identified, the system switched to the bridge state, in which the motors operated under the tool-specific parameters until the tool was removed, at which point the system reverted to the searching state to await the next tool.

\subsubsection{Master Tool Manipulator}

As seen in Fig. \ref{fig:wide_image}, the software pipeline starts with the Master Tool Manipulator (MTM), which publishes the pose from the user input onto a ROS topic to which the complete teleoperation pipeline subscribes, controlling the tool end effector. The MTM consists of two handles, MTML and MTMR, as seen in Fig. \ref{fig:wide_image}, that control the position and orientation of each tool end-effector. The position portion of the handle pose maps to the robotic manipulator, and the portion of the handle pose that defines orientation maps to the DASH motors. The controller calculates the inverse kinematics to determine the required motor positions. 

\subsection{Kinematics: Classic System}
The kinematics of the robot from the shoulder joint to the da Vinci Classic end effector tip position is divided into three sections: 1) the robot shoulder joint to robot end effector, 2) the robot end effector to the shaft origin, and 3) the shaft origin to the tool tip. From here, both the forward and inverse kinematics can be computed for the system.

\subsubsection{Robot Shoulder to Robot End Effector}
The forward kinematics of the humanoid arm are computed from the URDF as a product of $4\times4$ homogeneous transformation matrices:
\vspace{-3mm}

\begin{equation}
     ^{b}T_{e} = \prod_{i=1}^{n} T_i(q_i) \in \mathbb{R}^{4\times4}
\end{equation}

\noindent where $n$ is the number of joints and $q_i$ are the joint angles. The Unitree G1 has n = 7 joints from shoulder to wrist.

\subsubsection{Robot End Effector to Shaft Origin}

The rigid adapter transform from the robot end effector to the da Vinci roll pivot is:
\vspace{-2mm}

\begin{equation}
    ^{e}T_{s} = \begin{bmatrix} 
    \mathbf{I}_{3\times3} & \mathbf{d} \\ 
    \mathbf{0} & 1 
    \end{bmatrix} \in \mathbb{R}^{4\times4}
\end{equation}

\noindent where $\mathbf{d} = [0.073, 0, 0.098]^{\top}$ m   is the measured translation from the robot end effector to the roll pivot point on the da Vinci tool. 
\subsubsection{Shaft Origin to Tool Tip}
The da Vinci wrist kinematics are then described using the Modified  Denavit-Hartenberg (MDH) convention. As an example, the MDH parameters for the da Vinci Classic 8 mm Large Needle Driver are given in Table~\ref{tab:dh}. 

\begin{table}[t]
    \centering
    \setlength{\tabcolsep}{4pt}
    \caption{MDH Parameters -- 8 mm Large Needle Driver }
    \label{tab:dh}
    \begin{tabular}{lccccc}
        \toprule
        Frame & $\alpha_{i-1}$ (rad) & $a_{i-1}$ (m) & $\theta_i$ (rad) & $d_i$ (m) & Type \\
        \midrule
        Roll         & $0.0000$  & $0.0000$ & $\phi_{r}$           & $0.4162$ & Revolute \\
        Wrist Pitch  & $-1.5708$ & $0.0000$ & $\phi_{p} - 1.5708$ & $0.0000$ & Revolute \\
        Wrist Yaw    & $-1.5708$ & $0.0091$ & $\phi_{y} - 1.5708$   & $0.0000$ & Revolute \\
        End Effector & $-1.5708$ & $0.0000$ & $0.0000$                & $0.0102$ & Fixed    \\
        \bottomrule
    \end{tabular}
\end{table}

Applying MDH convention, the four wrist transforms are:
\vspace{-5mm}

\begin{equation}
    T_1(\phi_{r}) = \begin{bmatrix}
        c(\phi_{r}) & -s(\phi_{r}) & 0 & 0 \\
        s(\phi_{r}) & c(\phi_{r})  & 0 & 0 \\
        0     & 0      & 1 & l_{tool} \\
        0     & 0      & 0 & 1
    \end{bmatrix}
\end{equation}

\begin{equation}
    T_2(\phi_{p}) = \begin{bmatrix}
        c({\phi_p}) & -s({\phi_p}) & 0 & 0 \\
        0     & 0      & 1 & 0 \\
        -s({\phi_p}) & -c({\phi_p}) & 0 & 0 \\
        0     & 0      & 0 & 1
    \end{bmatrix}
\end{equation}

\begin{equation}
    T_3(\phi_{y}) = \begin{bmatrix}
        c({\phi_y}) & -s({\phi_y}) & 0 & l_{P2Y} \\
        0     & 0      & 1 & 0 \\
        -s({\phi_y}) & -c({\phi_y}) & 0 & 0 \\
        0     & 0      & 0 & 1
    \end{bmatrix}
\end{equation}
\vspace{-2mm}
\begin{equation}
    T_4 = \begin{bmatrix}
        1 & 0  & 0 & 0 \\
        0 & 0  & 1 & l_{Y2C} \\
        0 & -1 & 0 & 0 \\
        0 & 0  & 0 & 1
    \end{bmatrix}
\end{equation}

In these transforms, $l_{tool}$ is the tool-specific shaft length from the roll pivot to the pitch pivot point along the insertion axis, $l_{P2Y}$ is the pitch-to-yaw offset, and $l_{Y2C}$ is the yaw-to-tool-tip offset. All variables, $l_{tool}$, $l_{P2Y}$ and $l_{Y2C}$ are updated when a tool is scanned via the RFID board.
\vspace{-2mm}

\begin{equation}
    ^{s}T_{t} = T_1(\phi_{r}) \cdot T_2(\phi_{p}) \cdot T_3(\phi_{y}) \cdot T_4
\end{equation}

where $\phi_{r}$ is the roll angle, $\phi_{p}$ is the pitch angle, and $\phi_{y}$ is the yaw angle. 

\subsubsection{Full Forward Kinematics}

The complete forward kinematics from robot base to tool tip is:
\vspace{-2mm}

\begin{equation}
   ^{b}T_{t} = {}^{b}T_{e}(\mathbf{q}) \cdot {}^{e}T_{s} \cdot {}^{s}T_{t}(\boldsymbol{\phi})
\end{equation}

\subsection{Inverse Kinematics}

Given a desired tool tip pose $^{b}T_{t}^{*}$, the wrist joint angles are solved by isolating the wrist transform:

\vspace{-4mm}

\begin{gather}
    {}^{s}T_{t}^{*} = \left({}^{e}T_{s} \right)^{-1}
    \cdot \left( {}^{b}T_{e}(\mathbf{q}) \right)^{-1}
    \cdot {}^{b}T_{t}^{*} \\
        {}^{s}T_{t}^{*} = T_1(\phi_{r}^{*}) \cdot T_2(\phi_{p}^{*})
    \cdot T_3(\phi_{y}^{*}) \cdot T_4
\end{gather}

where $(\cdot)^{*}$ denotes a desired solution, ${}^{b}T_{t}^{*}$ is the
commanded tool tip pose expressed in the robot base frame, and $\mathbf{q}$
is the measured arm configuration.

\subsubsection{Remote Center of Motion}
The tool shaft must pivot about a defined fixed location, the remote center of motion (RCM), $\mathbf{p}_{rcm}$, at a commanded insertion depth, $h^{*}$. The shaft origin is represented by $\mathbf{s}(\mathbf{q})$ and the unit shaft axis is $\hat{\mathbf{u}}(\mathbf{q})$, which are both obtained from the forward kinematics. This requires the following equation:

\vspace{-2mm}

\begin{equation}
    \mathbf{s}(\mathbf{q}) + \left(L - h^{*}\right)\hat{\mathbf{u}}(\mathbf{q})
    = \mathbf{p}_{rcm}
    \label{eq:rcm}
\end{equation}
\vspace{-4mm}

The point a distance $L - h^{*}$ along the shaft must intersect with the RCM, where $L$ is the full shaft length from shaft origin to tool tip. The remaining length $h^{*}$ is the portion inserted through the port. 
Equation \eqref{eq:rcm} is not perfectly satisfied, causing the controller to drive the residual,
\vspace{-2mm}

\begin{equation}
    \mathbf{r}(\mathbf{q}) = \mathbf{s}(\mathbf{q})
    + \left(L - h^{*}\right)\hat{\mathbf{u}}(\mathbf{q}) - \mathbf{p}_{rcm}
\end{equation}
\vspace{-5mm}

to zero. This imposes three scalar constraints on seven joints, leaving four redundant degrees of freedom, which are resolved by a posture objective. During operation, the shaft must not move outside of a 10 mm diameter around the RCM. 

\subsubsection{Coupling Matrices}
Each da Vinci Classic tool has a unique coupling matrix \cite{dvrk} $\mathbf{M_1} \in \mathbb{R}^{4\times4}$, which maps actuator disk angles to joint space controlling four degrees of freedom: roll, pitch, yaw, and grip. The inverse coupling matrix $\mathbf{M_1}^{-1}$ is used to convert desired joint angles to motor disk commands:
\vspace{-2mm}

\begin{equation}
    \begin{bmatrix} \theta_{D1} \\ \theta_{D2} \\ \theta_{D3} \\ \theta_{D4} \end{bmatrix}
    = \mathbf{M_1}^{-1}
    \begin{bmatrix} \phi_{r} \\ \phi_{p} \\ \phi_{y} \\ \phi_{g} \end{bmatrix}
\end{equation}

where $\phi_{g}$ is the grip angle. The wrist joint angles $\phi_{r}, \phi_{p}, \phi_{y}$ are extracted from ${}^{s}T_{t}^{*}$ in the preceding section.

\subsection{Kinematics: Xi Tool}

Similar to the Classic tools, each Xi tool has a unique coupling matrix, $\mathbf{M_2}$, that describes the relationship between the disk rotation angles and the resulting end effector motions. While the Xi tools can have up to five DOF, the tools used in this work have only four, so the coupling matrix considers only four motor inputs, motors 1 to 4. The coupling matrix is defined as:
\vspace{-2mm}

\begin{equation}
    \mathbf{M_2} = \begin{bmatrix}
        m_{11} & 0      & 0      & 0      \\
        0      & m_{22} & 0      & 0      \\
        0      & m_{32} & m_{33} & m_{34} \\
        0      & 0      & m_{43} & m_{44}
    \end{bmatrix}
\end{equation}

where each nonzero element of $\mathbf{M_2}$ represents the angular coupling between a specific disk rotation and an end effector motion and is calculated as the slope of the corresponding tool angle with respect to the disk rotation angle. Thus, based on the motor IDs shown in Fig. 4, $m_{11}$ and $m_{22}$ describe the roll and pitch responses to motors 1 and 2, respectively. The coefficient $m_{32}$ represents the yaw induced by motor 2 during pitch motion, while $m_{33}$ and $m_{34}$ describe the yaw responses to motors 3 and 4. The coefficients $m_{43}$ and $m_{44}$ describe the grip responses to motors 3 and 4. These coefficients provide the basis for converting between disk rotation commands and the resulting tool motions.

\subsection{Teleoperation}
To direct the robot, the user commanded the arms using the MTM as shown in Fig. \ref{fig:wide_image}: the desired tool tip pose was then mapped from the MTM handle pose, and the consequent joint configuration was solved according to the given RCM. The RCM was placed at [0.4, -0.15, 0.05] m for the right arm and [0.4, 0.15, 0.05] m for the left arm, as expressed in the robot base frame. 

\section{Experiments}
\label{sec:experiments}

\subsection{Adapter Validation}


Table~\ref{tab:results} summarizes the results for both adapters. Each adapter, with the heaviest instrument inserted, meets the 0.5~kg mass requirement (Table~\ref{tab:dr}), and both have onboard motor latencies below 6~ms. Tool identification and zeroing were evaluated by repeatedly inserting four tools into DASH~C and two into DASH~Xi. A trial was successful if the correct tool-specific kinematics were loaded and the tool visibly returned to its zero position.  Tool exchange time was measured from first contact with the installed tool to full seating of the next tool.

\begin{table}[t]
    \centering
    \small
    \setlength{\tabcolsep}{3pt}
    \caption{DASH Experimental Results}
    \label{tab:results}
    \begin{tabular}{lcc}
        \toprule
        Test & DASH C & DASH Xi \\
        \midrule
        Mass (kg)                    & $0.439$                & $0.463$ \\
        Last-Motor Latency$^a$ (ms)  & $5.07 \pm 0.04$   & $5.01 \pm 0.05$ \\
        Tool ID \& Zeroing$^b$          & $40/40$           & $20/20$ \\
        Tool Exchange Time$^c$ (s)   & $4.74 \pm 1.15$   & $3.39 \pm 0.51$ \\
        \bottomrule
        \multicolumn{3}{l}{\footnotesize $^{a}$5 trials $\times$ 20 pings; Xi motor 5 unused due to 4-DOF tools}\\ 
        \multicolumn{3}{l}{\footnotesize $^{b}$4 (C) or 2 (Xi) tools $\times$ 10 insertions each} \\
        \multicolumn{3}{l}{\footnotesize $^{c}$10 trials; mean $\pm$ SD} \\
    \end{tabular}
\end{table}

\subsection{Xi Tool: Coupling Matrix}
\label{subsec:coupling_matrix}

Since the coupling matrices for the Xi tools were not publicly available, each matrix was experimentally measured. The motor zero positions were first defined with the end effector in the straight configuration, and the X–Y–Z axes were established using the right-hand rule. The open grip angle was defined as positive. Tool motion was measured in 3D using Kinovea from multiple camera orientations and correlated with the corresponding motor angles recorded by the control code. The grip coefficients $m_{43}$ and $m_{44}$ were obtained from the yaw coefficients $m_{33}$ and $m_{34}$ by multiplying their magnitudes by two, with the sign of $m_{44}$ reversed to account for the opposing motor directions.
 
\begin{figure}[t]
  \centering \includegraphics[trim={0 0 0 0 cm}, clip, width=0.94\linewidth]{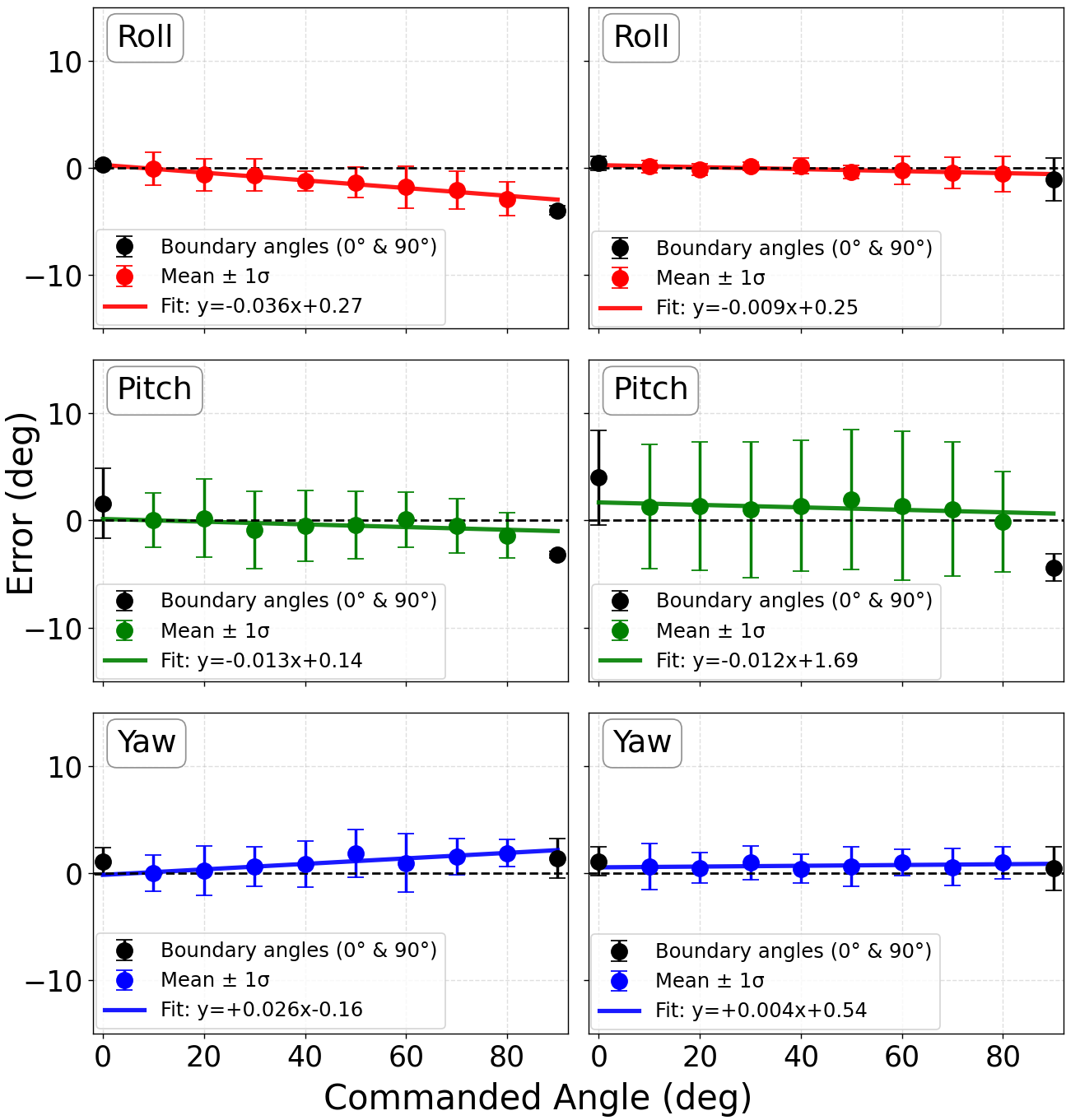}
  \caption{Error vs. commanded angle for the 8 mm Mega SutureCut Needle Driver (left) and 8 mm Tenaculum Forceps (right) in roll, pitch, and yaw for DASH Xi. The mean angular error and ±1 standard deviation are shown for commanded angles from 0° to 90°. Best-fit lines were calculated excluding the data points at 0° and 90°.}
  \label{fig:backlash}
  \vspace{-4mm}
\end{figure}

 To characterize the different sources of error, two 8 mm da Vinci Xi instruments, the Tenaculum Forceps and Mega SutureCut Needle Driver, were tested from 0° to 90° and back to 0° in both directions, with measurements at 10° increments. As seen in Fig. \ref{fig:backlash}, the data provides insight into the contributions of measurement error, adapter backlash, coupling-matrix error, zero-position offset, and tool backlash. The best-fit lines exclude the 0° and 90° boundary points, where motion is limited to a single direction and backlash causes the error to deviate systematically from the fit. A negative slope indicates that the coupling matrix underestimates angular magnitude, and vice versa; the intercept reflects zero-position offset. The error bars in roll are primarily attributed to adapter backlash, as the Xi roll joint uses geared transmission with lower backlash than the cable-driven joints. In contrast, the larger variation observed in pitch for the Tenaculum Forceps suggests additional backlash within the tool. Near-zero slopes indicate acceptable coupling-matrix accuracy, with further calibration possible using the same procedure.

\begin{figure}[t]

    \centering
    \includegraphics[trim={0cm 3cm 17.5cm 6.5cm}, clip, width=0.98\linewidth]{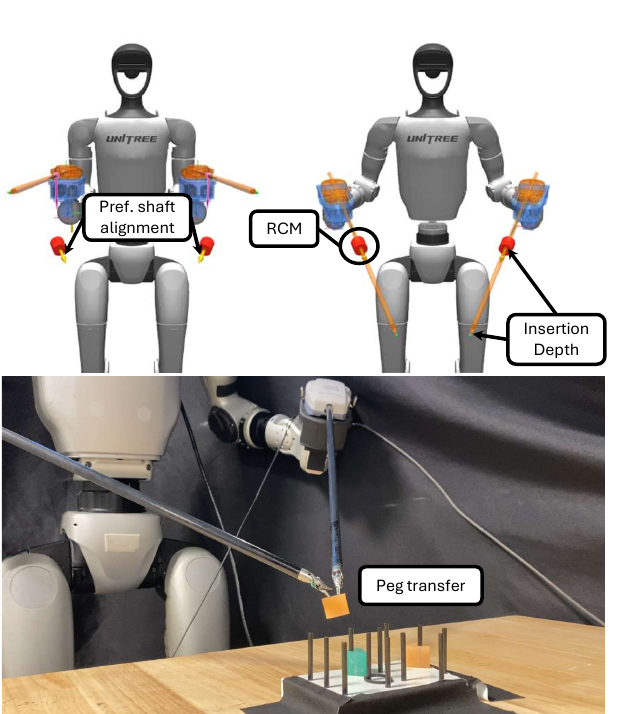}
    \caption{Initialization of humanoid platform with DASH. The top left and right panels show the neutral and initialized poses, respectively. The RCM position, insertion depth, and instrument orientation are predefined to initialize the robot configuration before operation. The bottom panel displays the humanoid being teleoperated to complete a peg transfer task.}
    \label{fig:sim2real}
\end{figure}
 
\subsection{Simulation of System}

A simulation of the humanoid robot with DASH was developed using ground-truth system dimensions, providing a controlled environment for the development and validation of RCM control in the humanoid setting. This simulation supported system validation across the full control pipeline, from RCM initialization to MTM-based teleoperation.
As seen in Fig. \ref{fig:sim2real}, the robot starts at the neutral pose and places the tool with respect to the desired RCM. The final arm position was determined based on insertion depth and RCM location in the base frame, with the insertion depth as the distance from the tip of the end effector to intersection with the RCM. Once the pose initialization was completed, the MTM was used to control the end effector position and orientation. The simulation validated that the robot can orient the end effector about the RCM constraint.

\subsection{Teleoperative System Validation}
In order to validate that the humanoid platform with DASH C can sufficiently perform tasks, as demonstrated in Fig. \ref{fig:sim2real}, the latency of the system and the motion accuracy at the tip and RCM of the system was recorded.

\subsubsection{Full System Latency}
 The end to end system latency was measured with an external camera at 120 fps across three categories: moving the MTM up/down, left/right, and forward/backwards. Across the three categories the full system latency was 25.8 ± 18.2 ms, 12.0 ± 4.4 ms, and 19.4 ± 13.7 ms respectively. Across all latency measurements, the total latency is 20.3 ± 15.7 ms, substantially lower than the previously reported latencies at 156 ms~\cite{liang2026}.



\subsubsection{Command-Execution Tracking}
\begin{figure}[t]
    \centering
    \includegraphics[trim={1.5cm 2.5cm 0 3.5cm}, clip, width=0.98\linewidth]{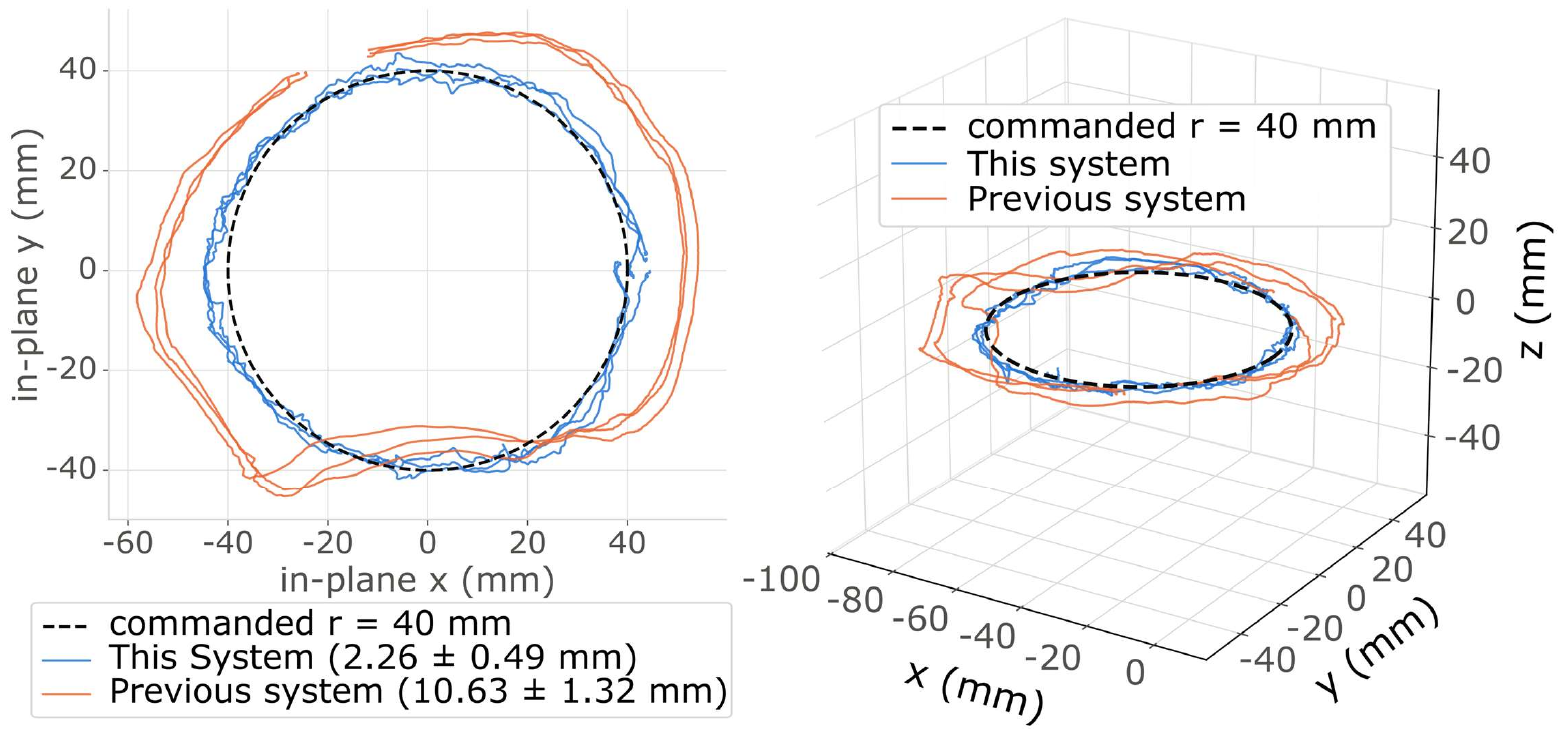}%
    \caption{Comparison of commanded tool tip trajectory vs executed trajectory with DASH and the previous system. The left and right graphs show the in-plane 2D radial RMS error and 3D view, respectively, with the DASH trajectory substantially more accurate than the previous system.}
    \label{fig:traj}
\end{figure}

The motion accuracy for DASH was compared to the previous system, which utilized Livsmed tools mounted to the humanoid platform and a constrained location for the RCM. DASH enables the system to have a defined RCM, removing the need for a physically constrained RCM. The system was commanded to make a circle with a radius of 40 mm at the tip, showcasing the ability for the constrained RCM to still have precise control. As seen in Fig. \ref{fig:traj}, this system had a resulting radius of 41.45 ± 0.30 mm over three trials, with the previous system having a resulting radius of 49.83 ± 1.09 mm with the same 40 mm commanded radius. The out of plane error for the system with DASH was 1.50 ± 0.61 mm, and the previous system had an error of 1.23 ± 0.14 mm. The lower out of plane error for the previous system is likely due to there being a physical RCM constraint, limiting vibrations and physically preventing the system from going far out of plane, whereas the DASH system has a remote RCM constraint with no physical constraint.

\subsubsection{RCM Convergence}
\begin{figure}[t]
    \centering
    \includegraphics[trim={1cm 0 0 2.5cm}, clip, width=0.92\linewidth]{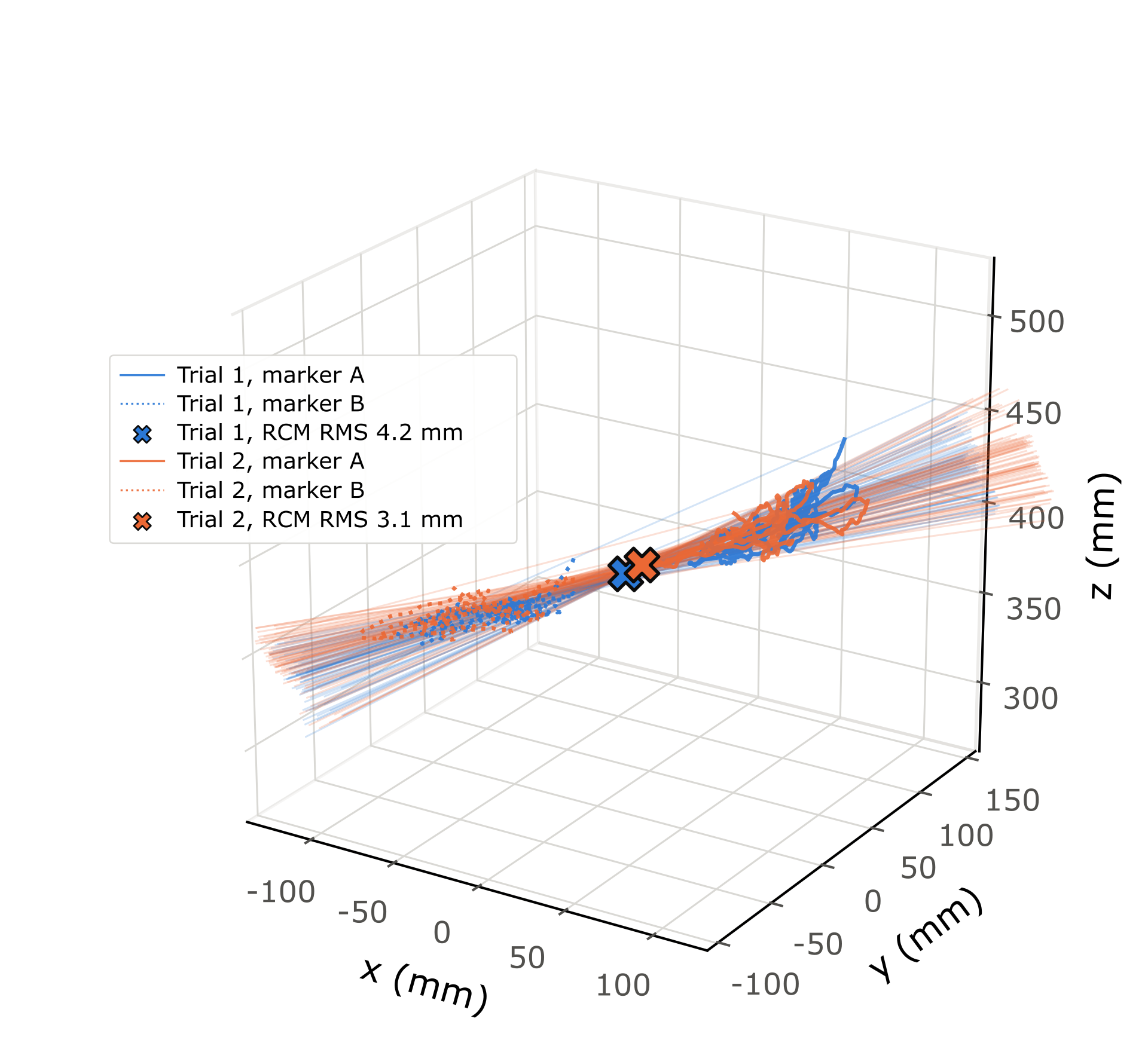}%
    \caption{RCM convergence across two trials. Optitrack markers placed on either side of the RCM tracked the shaft's motion over random trajectories which was used to determine the converged RCM and pivot error.}
    \vspace{-5mm}
    \label{fig:rcm}
\end{figure}

To determine the precision of the calculated RCM pose, two trials were conducted using an OptiTrack camera and two OptiTrack markers. Similarly to \cite{rcm_conv}, the trajectories of the markers were tracked, putting infinite lines through the points to determine the convergence of the RCM, as seen in Fig. \ref{fig:rcm}. The two estimated RCM positions are 15.79 mm apart, compared to a 410 mm shaft. Each respective RCM had a pivot error of 3.37 ± 2.44 mm and 2.64 ± 1.54 mm, which was within the 10 mm bound requirement.

\section{Discussion and Conclusion}
\vspace{-2mm}

In this paper, we demonstrate that surgical-grade da Vinci Classic and Xi instruments can be reliably decoupled from the proprietary da Vinci platform and operated on general-purpose robots via DASH. This decoupling is seen in Fig. \ref{fig:coverFig}, where a general purpose humanoid robot uses both DASH C and DASH Xi. These combined results show that DASH performs reliably and improves on the previous system which used passively linked laparoscopic tools. However, this work has not validated DASH under a full surgical task, sterilization constraints, or with a surgeon operating the MTM, all of which are necessary before any clinical relevance can be claimed.

This work will benefit both the healthcare and robotics communities in meaningful ways. For healthcare, demonstrating that standardized da Vinci instruments can be reliably interchanged on compact, more versatile platforms opens a more realistic path toward robotic assistance in hospital settings that do not currently have access to robotic surgical platforms. For the robotics community, the ability to mount and interchange clinical-grade surgical instruments on widely available platforms such as serial-link and humanoid systems significantly lowers the barrier to surgical robotics research, enabling experimentation and development that does not depend on access to a full da Vinci installation or the da Vinci Research Kit. 

\section{Acknowledgments}
\vspace{-1mm}

In this paper, Claude Code was used to help implement described code and scripts to graph the data. Additionally, Claude and ChatGPT were used to give feedback on writing clarity and continuity, but did not generate any of the prose. 

\vspace{-3mm}

\bibliographystyle{ieeetr}
\bibliography{refs}
\vspace{-3mm}

\newpage

\end{document}